How to integrate cloud service, data analytic and machine learning technique to reduce cyber risks associated with the modern cloud based infrastructure

Upakar Bhatta
*Marymount University*
Arlington, VA, USA
upakarb@gmail.com

In today's dynamic and competitive digital era, companies are leveraging cloud technology, machine learning, and data visualization techniques to reinvent their business processes. The combination of cloud technology, machine learning, and data visualization techniques allows hybrid enterprise networks to hold massive volumes of data and provide employees and customers easy access to these cloud data. These massive collections of complex data sets are facing security challenges. While cloud platforms are more vulnerable to security threats and traditional security technologies are unable to cope with the rapid data explosion in cloud platforms, machine learning powered security solutions and data visualization techniques are playing instrumental roles in detecting security threat, data breaches, and automatic finding software vulnerabilities. The purpose of this paper is to present some of the widely used cloud services, machine learning techniques and data visualization approach and demonstrate how to integrate cloud service, data analytic and machine learning techniques that can be used to detect and reduce cyber risks associated with the modern cloud based infrastructure. In this paper I applied the machine learning supervised classifier to design a model based on well-known UNSW-NB15 dataset to predict the network behavior metrics and demonstrated how data analytics techniques can be integrated to visualize network traffics.

I.  INTRODUCTION

Throughout the last decade, the numbers of malicious incidents have dramatically increased causing a significant impact on the organization infrastructure [1]. Cyber-attacks and security threats have dramatically increased in emerging technologies such as Cloud and Internet of Things (IoT). During the Covid-19 pandemic, IT professionals see an increase in cyber-attacks such as distributed denial-of-service (DDoS) that targets normal network traffic by injecting malicious network traffic and thus disrupting the organization's system. Several research have been done previously regarding how to monitor and predict network attacks using a machine learning approach to reduce cyber-attacks and security threats. The enterprise security analysis has been highlighted [2]. The investigation of a feasibility study of applying machine learning techniques and overview of state of art machine learning technique for network and cloud security has been outlined [3]. However, previous research doesn't provide enough detail regarding implementation of machine learning technique to predicate malicious activates.

This research study explores the cloud services, data visualization tools, and machine learning techniques and demonstrates these services can be applied to select the network parameter to proactively detect and predict the various types of network attacks.

A.  Proposed Idea

The purpose of this research study is to explore cloud services, machine learning and data visualization techniques and learn how to integrate cloud visualization and machine learning capable service to visualize network anomaly traffic.

B.  Problem Statement

Rodney Alexander highlighted the enterprise security analysis in his paper "Using Linear Regression Analysis and Defense in Depth to Protect Data Breach and Network attacks during the Global Corona Pandemic" [3]. According to Thales Security, 67 percent of enterprises across the world have at least one incident of a major data breach or network attack annually [4]. The general problem is to understand the challenge of machine learning implementation. The specific problem is to understand how effective cloud services, machine learning techniques, and data analytics tools are at detecting and visualizing network attacks and the vulnerabilities. The research gap is that there is limited research on how new emerging applications are integrating cloud services, machine learning and data visualization techniques to focus on tracking and measuring cybersecurity breaches.

II.  PRE-REQUSITE KNOWLEDGE

A.  CLOUD COMPUTING

Cloud computing is the process of connecting virtual resources such as compute, storage, networks, and databases over the internet in order to share the data. Cloud computing is one of the fastest growing technologies that uses computing, storage, networking, and database tools to share virtual resources through the use of the Internet. However, there multiple challenges that the organization has to face in regard to data security and privacy when it adopts the cloud computing infrastructure [19].

The following are the three cloud computing models that the most of the today's organizations are leveraging:

Infrastructure as a Service (IaaS): It offers the virtualized computing resources over the internet and provides the highest level of flexibility and management control over the infrastructure compared to other models.

Platform as a Service (PaaS): It provides the customer with an underlying infrastructure to deploy their own code.

Software as a Service (SaaS): It offers a complete software packages over the internet on a subscription basis.

B. BIG DATA ANALYTICS

Big data analytics is a process of examining large data sets to help business understand hidden patterns, and insights. Companies can leverage analytical tools and technologies to study big data deeply to understand the hidden business insights [20]. Data velocity, data variety, and data volume are the key terms used in big data analytics process. Data velocity is basically the speed at which the data is collected. There are more than 3.5 billion Google searches per day [21]. Data variety refers to various types of data such as structured, unstructured, and semi-structured that might be captured during data analytic process. Big data analytics deals with huge volumes of data that may come from different sources. For instance, the average mobile traffic was 6.2 Exabyte per month; Netflix has over 86 million members globally [21].

C. MACHINE LEARNING

Machine learning is a mathematical technique which basically focuses on building algorithms and statistical models to enable computers to make predictions by analyzing the hidden pattern on the dataset. Machine learning has introduced a new approach to train the algorithms based on real-time datasets to identify specific patterns and anomalies in network traffic [22]. There are three different categories of machine learning that are supervised, unsupervised, and reinforcement learning.

In supervised machine learning technique, the algorithm is trained based on the labeled dataset. Supervised learning is also divided into Classification and Regression [23]. In classification tasks, the output is basically a category whereas in regression tasks the goal is to predict a continuous value.

In unsupervised learning technique, the algorithm is trained based on unlabeled data. Since unsupervised learning techniques deals with unlabeled data, its learning process is complex. Unsupervised learning being so complicated, and complex is used quite fewer times than supervised learning [24].

In reinforcement learning technique, system learns through the trial and error process and gets feedback in the form of rewards and punishment. Through trial and error, the algorithm determines which steps result in the highest rewards [25].

## III. METHODOLOGY

In this paper, machine learning algorithm and the ML capable cloud services have been explored to perform and analyze network behavior using UNSW-NB15 dataset. I demonstrated four machine learning algorithm to predict the accuracy of network traffic. The four machine learning algorithm that are explored in this paper are Naïve Bayes, Ada Boost, K-Nearest Neighbors, and Random Forest Classifier.

To analyze the network behavior using machine learning technique, the following approach has been implemented:

- Acquiring dataset, Data preprocessing , Split the dataset into training and testing segments, Build ML Models, Train ML Models, Predict the output, Find the accuracy, Evaluate the model

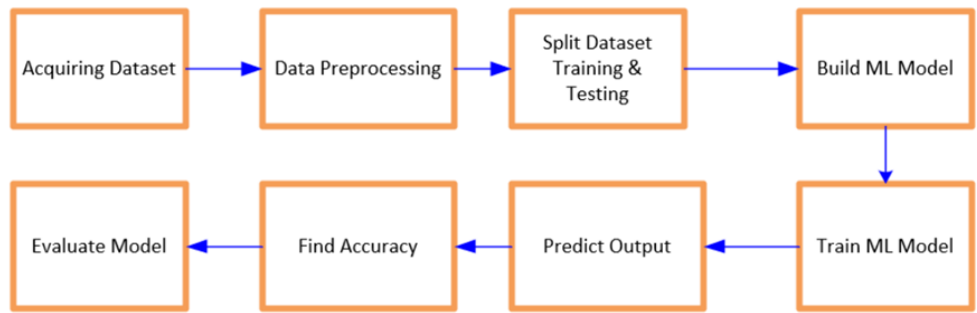

Figure 1: Machine learning implementation steps

To visualize network traffics using cloud services, the following approach has been implemented:

- Leveraging Cloud storage to host the dataset
- Leveraging cloud processing tool that can be used to process the data and store the process data back to the target storage
- Leveraging query interactive tool to interact data resides in target storage
- Leveraging cloud visualization tools to display Queried data from cloud storage.

A. Data Analysis

This research used experimental methodology to use the UNSW-NB15 dataset to categorize the normal and abnormal traffics based on the protocol, tcp flag, and port number. The UNSW-NB15 dataset [5] was published in 2015 which basically includes nine different attack types and 49 features. The experiment research method selected for this study is exploratory where the features extracted from the collected dataset were used to analyze the network traffic accuracy.

The initial step involves importing the dataset files, data understanding and cleanup such as source ip address, source port, destination ip address, destination port, protocol, bytes in, bytes out, packets in, packets out, tcp flags, flow duration, label, and attack. Usually, the data sets contain various important and redundant features which may affect the accuracy [6]. Therefore, it is important to select the appropriate feature to reduce overfitting and enhance the model accuracy. The UNSW-NB15 dataset contains 49 features. However, all 49 features are not important or relevant to the class labels that are used to categorize normal and abnormal network traffic. Since some features are specific to the computing infrastructure such as source IP address and destination IP address, they do not possess relevant information for anomaly detection purposes. The important input features that are selected during the exploratory analysis are L7 protocol, L4 destination port, and the TCP flags. In order for better accuracy, it is important that the highly correlated features are dropped properly. Therefore, the explanatory input features such as Source IP address, destination IP address, destination packets are eliminated. The label is binary and the attack_cat feature which contains the names of attack categories are defined as normal and abnormal network traffics.

B. Libraries, Preprocessing and feature engineering before training of ML models

Python based libraries that are used to perform the machine learning activities in this paper are Scikit-learn for data analysis purpose. Labels are encoded using Label Encoder to convert categorical data into a numerical format suitable for machine learning algorithms. This step ensures uniformity in handling categorical variables, supporting accurate model training. I performed exploratory data analysis on the UNSW-NB15 using the most commonly used visual representation techniques that is histogram visual representation of the distribution of numerical data. This visualization technique is the graphical representation of data that allows complex data to be communicated in a clear manner and help identify the normal and abnormal network traffic. Most cybersecurity professional used this technique to visually describe the network traffic pattern.

```python
import numpy as np
import pandas as pd
import matplotlib.pyplot as plt
from imblearn.over_sampling import SMOTE
from sklearn.neighbors import KNeighborsClassifier
from sklearn.model_selection import train_test_split
from sklearn.ensemble import RandomForestClassifier
from sklearn.metrics import accuracy_score, classification_report
from sklearn.preprocessing import StandardScaler
from sklearn.ensemble import AdaBoostClassifier
from sklearn.tree import DecisionTreeClassifier
from sklearn.preprocessing import LabelEncoder
from sklearn.naive_bayes import GaussianNB
from sklearn.metrics import f1_score
#from sklearn.preprocessing import StandardScaler
dataframe = pd.read_csv('/Users/upakarbhatta/Desktop/UNSW-NB15.csv')
#print(dataframe)
label_encoder = LabelEncoder()
dataframe['L4_DST_PORT'] = label_encoder.fit_transform(dataframe['L4_DST_PORT'])
dataframe['L7_PROTO'] = label_encoder.fit_transform(dataframe['L7_PROTO'])
dataframe['TCP_FLAGS'] = label_encoder.fit_transform(dataframe['TCP_FLAGS'])
print(dataframe.columns)
print(dataframe.head())

x = dataframe[['L4_DST_PORT','L7_PROTO','TCP_FLAGS']]
y = dataframe['Label']
```

Figure 2: Libraries used

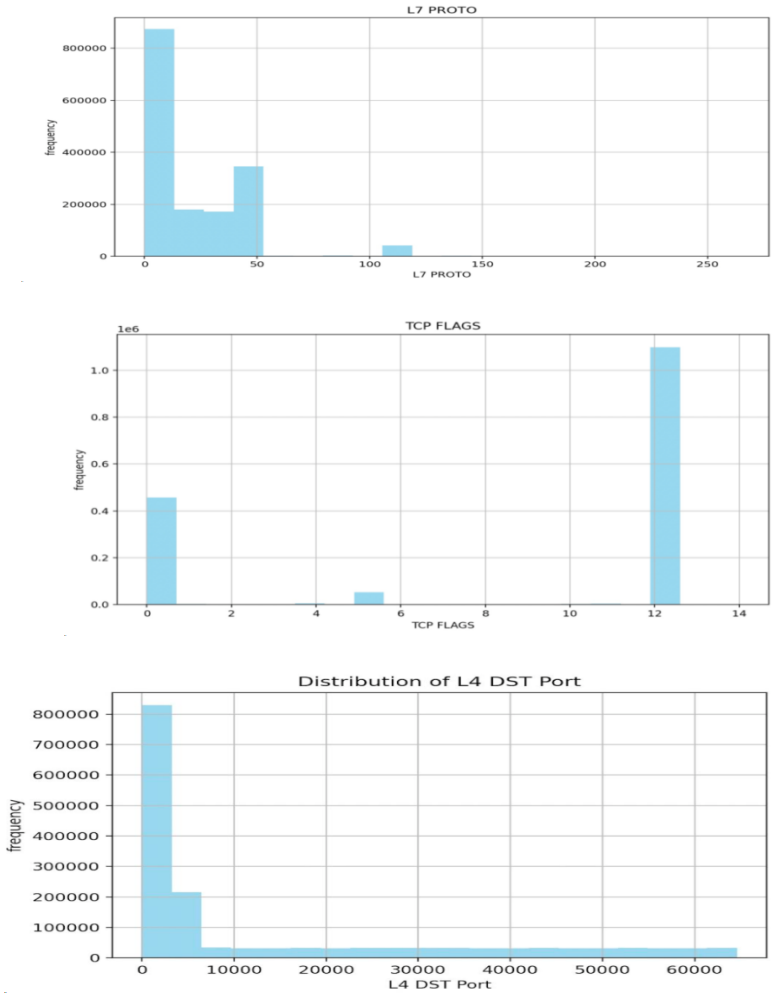

Figure 3: Plot of important features

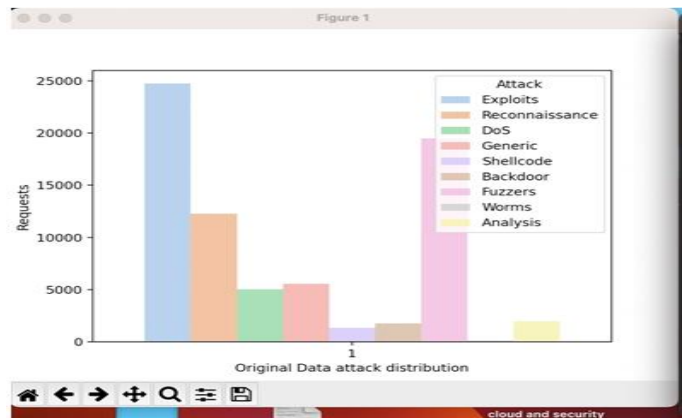

Figure 3: Plots of network attacks

Furthermore, for exploratory analysis, correlation heatmap is used to identify or check if the variables have a strong correlation to one another and remove highly correlated data. This is one of the important steps when using the dataset to build machine learning models. A correlation heatmap represents a visual graphic that demonstrates how the variables defined in the dataset are correlated to one another. -1 signifies zero correlation, while 1 signifies a perfect correlation. If the inputs are not independent of each other than the model cannot distinguish the significance of inputs in predicting the target outcome. In order for better accuracy, highly correlated features are dropped and features with low correlation with each other are kept.

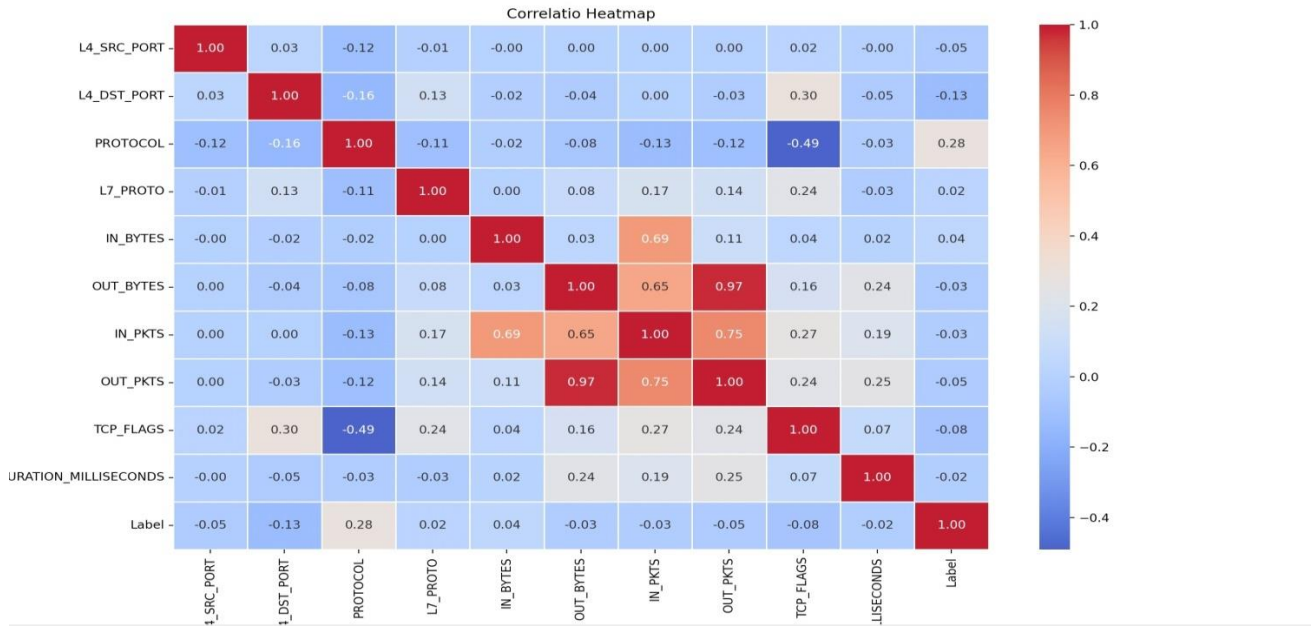

Figure 4: Correlation Heatmap

Exploratory data analysis can be beneficial to analyze the patterns in the data and helps select the appropriate features that would potentially improve the accuracy of machine learning model. Exploratory data analysis is useful to understand the distribution, relationships, and patterns in the data and enables the selection and preprocessing of appropriate features to improve the accuracy of the machine learning model [7]. For exploratory analysis, correlation heatmap is used in this paper to identify and remove highly correlated values.

C. Splitting the dataset into the training and testing segments

The preprocessed dataset is split into training and testing sets, with an 80-20 ratio. Splitting the dataset to ensure proper division between the training and testing segment is vital in predicting the output with better accuracy. This division allows for robust model training for the larger portion of the data while reserving a separate portion for evaluating model generalization and performance on unseen data during testing.

To deal with the imbalance dataset associated with the NB15, Synthetic Minority Over-sampling Technique (SMOTE) has been used to address class imbalance in classification problems. This technique is used to solve the class imbalance problem in supervised machine learning model.

D. Model Development

Machine learning model selection plays a vital role as it directly impacts the accuracy and performance of machine learning system. It is very important to meticulously choose machine learning methods and datasets based on their respective characteristics [8]. The first model demonstrated in this paper was the Naive Bayes classification model, which provided predictions based on the inherent probabilistic relationships between features. This model serves as a baseline for comparison with more complex algorithms. The Naïve Bayes (NB) classifier belongs to member of probabilistic family of classifiers based on Bayes' Theory and the main feature of this classifier is the assumption that all variables are conditionally independent which is the reason for calling it 'Naïve' [9]. Naïve Bayes classifier is very effective when dealing with large datasets like UNSW-NB15.

The second model employed on the UNSW-NB15 network traffic dataset was the Random Classifier classification model, an ensemble of decision trees that can handle a large number of input variables. Random Forest is usually applied to classification and regression issues that combines multiple models for robust performance. The Random Forest technique reduces overfitting and increases prediction accuracy by combining numerous decision trees [10]. Previous research shows the Random Forest classification model is more accurate than other classification models with the accuracy of 97.49%. In [11], the performance of various classification machine learning models has been implemented using Apache Spark and compared with related research based on the UNSW_NB15 dataset. This study states the accuracy of Random Forest model is much higher compared to Naive Bayes. The main advantage of Random Forest in comparison to individual decision trees is that it is less prone to overfitting [12].

The third model employed on the UNSW-NB15 network traffic dataset was the K-nearest neighbors (KNN), a supervised machine learning technique that is used to categorize similar malicious activities based on various distance functions. kNN is the basis of many of the clustering algorithms that are in use today [13]. KNN techniques is basically used to profile malware behaviors and to categorize them into malicious and normal classes [14]. KNN classifier technique was applied to categorize malicious and normal activities based on the new features. Although KNN is a strong anomaly detection technique, these techniques need a large storage space for the classification of high-speed traffic data and considered as very time consuming [15,16]. In this paper, KNN is basically used as a baseline algorithm for comparison with other models.

The fourth model employed on the UNSW-NB15 network traffic dataset was Adaboost classifier,

an ensemble learning technique used for both classification and regression issues. AdaBoost approach works by creating a strong classifier from a number of weak classifiers [17]. This algorithm basically good for moderate-sized dataset that works by combining multiple weak decision trees to create a strong learner.

E. Results and comparison

| ML Model | Accuracy | Metrics | Score |
|---|---|---|---|
| Naïve Bayes Classifier | 0.930784538419 | Precision | 0.92 |
| | | Recall | 0.93 |
| | | F1-score | 0.92 |
| AdaBoost Classifier | 0.954889348908 | Precision | 0.91 |
| | | Recall | 0.95 |
| | | F1-score | 0.93 |
| Random Forest Classifier | 0.954889348908 | Precision | 0.93 |
| | | Recall | 0.95 |
| | | F1-score | 0.93 |
| K-Nearest Neighbors Classifier | 0.688556607028 | Precision | 0.90 |
| | | Recall | 0.69 |
| | | F1-score | 0.78 |

Table 1: Machine learning model comparison.

**Key Considerations**

1. **Precision**:
   - **High Precision**: Few false positives which indicates that the obtained result is close to true value.
   - **Low Precision**: Many false positives which indicate that the obtained result varies significantly.

2. **Recall**:
   - **High Recall**: Few false negatives which indicate that the model correctly identifies the true value.
   - **Low Recall**: Many false negatives which indicate that the model fails to correctly identify the true value.
   - 
3. **F1 Score**:
   - **High F1 Score**: Indicates a good balance between precision and recall. It indicates that the model correctly identifies positive instances (high recall) with few false positive (high precision).

**Interpretation of Metric Scores:**

1. **Precision (0.91-0.94)**: ML model with a high precision with this range means few false positive and correct 94% of the time. A high precision rate of 94% is suitable for application if false positive is highly undesirable.
2. **Recall (0.69-0.95)**: ML model with high recall such as 0.95 identifies 95% of all actual positive instances. The lower range basically suggests that 69% is missing 31% of positive instances.
3. **F1 Score (0.78-0.93)**: The F1 score ranges 0 to 1 with 1 being the best and 0 being the worst. The F1 score is basically considered as a balance between precision and recall, and a score of high range 0.93 indicates that the model performs very well in balancing precision and recall. The lower range suggests that there is room for improvement.

F. MODEL ANALYSIS AND VISUALIZATION

Random Forest and AdaBoost classifier both provided better accuracy. With precision of 0.93, recall of 0.95, and an F1 score of 0.93, Random Forest's performance indicates the model's performance is excellent.

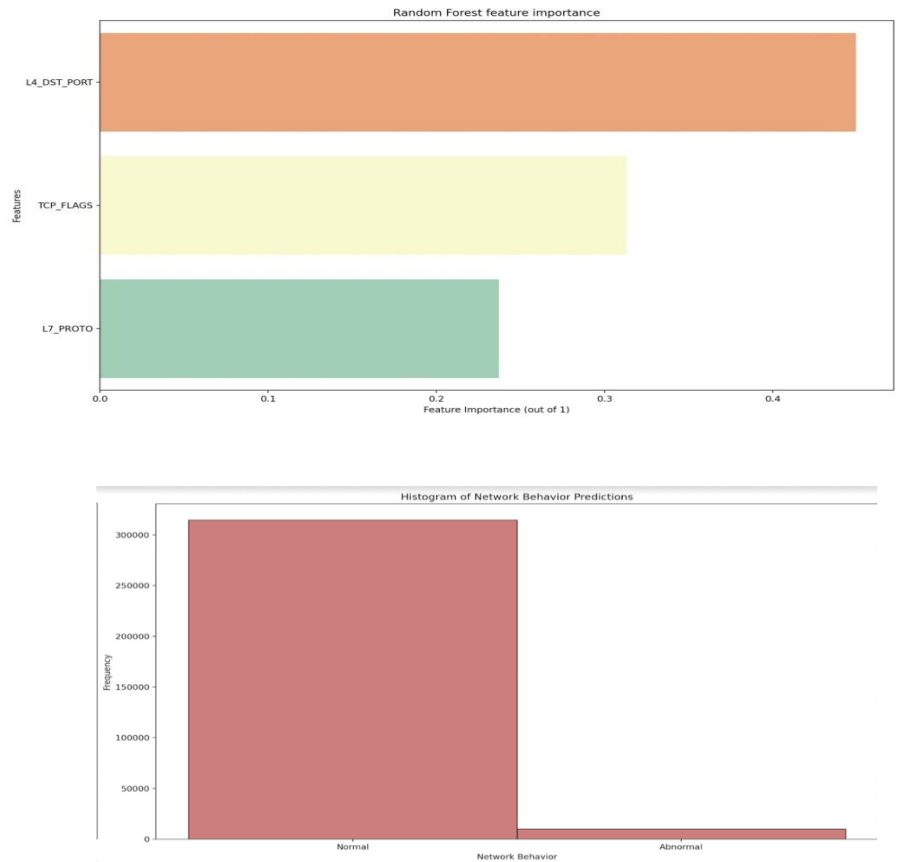

Figure 5. Random Forest selected feature and output.

Random Forest is chosen for anomaly detection in this research paper for following reasons:

- Random forest is known for its high accuracy, and it works well with the network feature
- Random Forest ensemble learning method is beneficial for this research as it improves performance compared to other model to detect anomalies effectively
- Compare of other models, Random Forest can handle both numerical and categorical features without any extensive preprocessing to analyze diverse datasets that I used in this research
- Random Forest can be tuned through cross validation to optimize performance for anomaly detection activities and to achieve the appropriate balance between specific network features in anomaly detection.

G. ML CAPABLE CLOUD SERVICES INTEGRATION

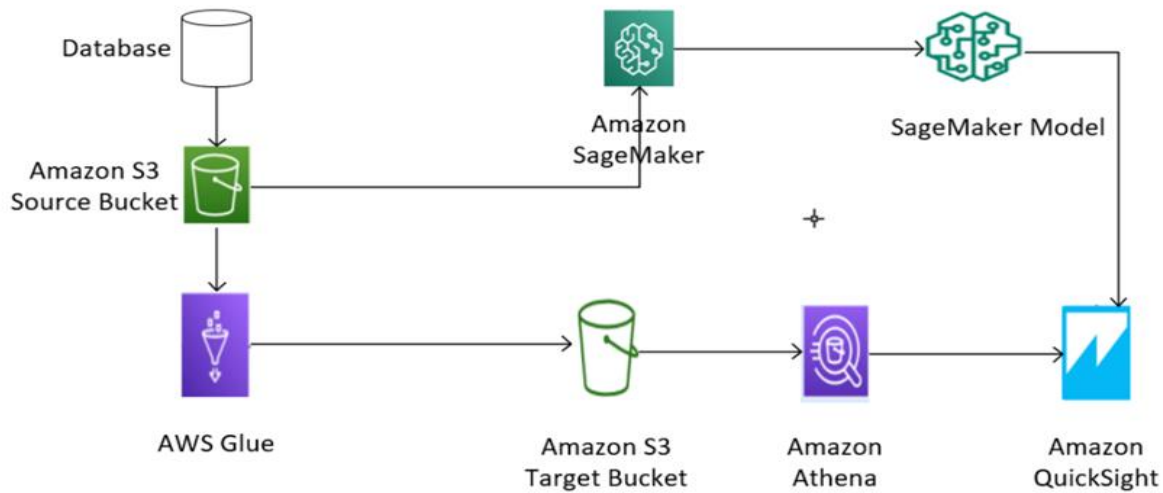

Figure 6. Proposed architecture using AWS (Amazon Web Services) Managed service

AWS serverless capabilities such as S3, Athena, QuickSight, and SageMaker are critical for modern enterprises to modernize an analytics workflow [18]. The proposed architecture that uses the AWS managed services are Glue to prepare data for analysis, SageMaker to build and train ML model, Athena to interact data store in S3, QuickSight to create visualization and dashboards

This architecture plan outlines the cloud services that can be used to perform, analyze, and visualize network behavior using NB15 dataset on Amazon Web Services (AWS).

**Architectural workflow:**

- Upload the dataset to Amazon S3

- Preprocess the dataset using AWS Glue and put the preprocessed dataset into another S3 bucket
- Setting up an AWS Glue to preprocess the data stored in S3. The NB15 dataset will be reviewed to understand its structure and relevance to network behavior analysis. Python with Pandas will be used to manipulate the data, which involved filtering, cleaning, and aggregating to suit the project's objectives
- Querying Data with Amazon Athena. Amazon Athena, an interactive query service, is used to run SQL queries against the preprocessed data in S3. This step involved setting up Athena, creating a database, and defining a table that maps to the data structure of the NB15 dataset. Queries are written to get insights into network behavior.
- Incorporate SageMaker that provides a fully managed ML service to build and train ML model
- Data Visualization with Amazon QuickSight

Finally, Amazon QuickSight, a business intelligence service, was used to visualize the queried data. The service was connected to Athena as a data source, and various visualizations were created to represent different aspects of the network data, facilitating easy interpretation and insight extraction.

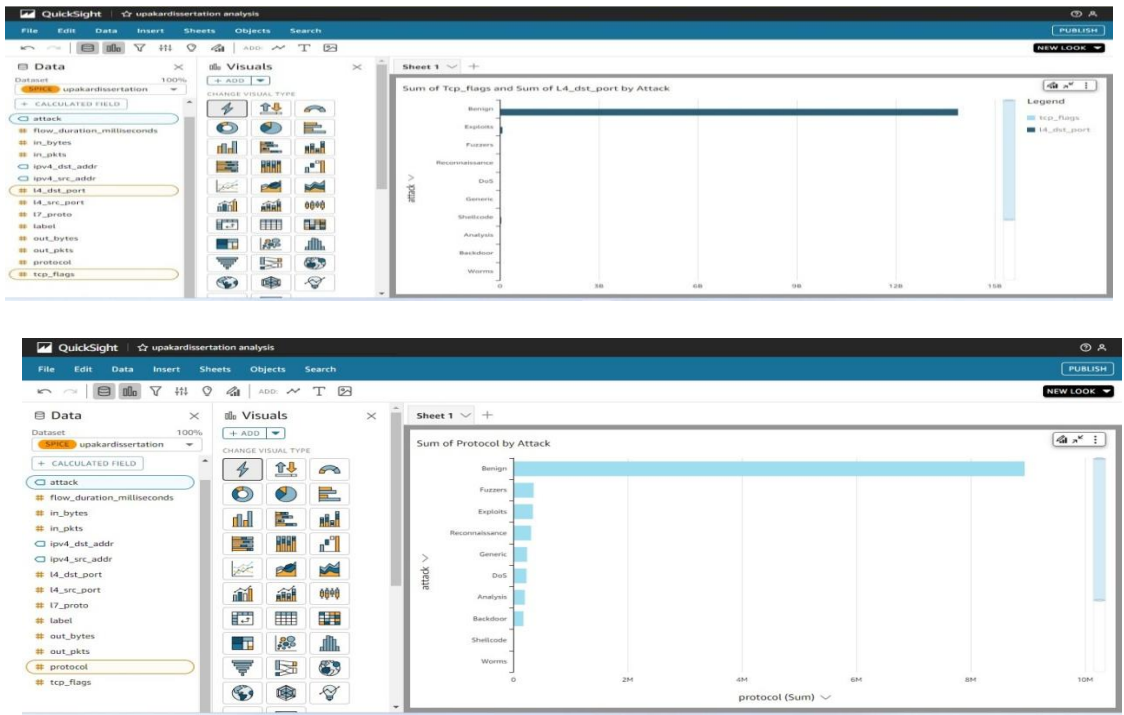

Figure 6. Visualization using Amazon QuickSight

The above figure shows the interactive dashboards in QuickSight to analyze network threats. I leverage Amazon QuickSight to create interactive dashboard based on well-known UNSW-NB15 dataset. Amazon QuickSight dashboards are very helpful in terms of analyzing network traffic effectively.

## IV. CONCLUSION

This paper explored the cloud services and machine learning techniques and discussed how cloud services, machine learning techniques and data visualization can be integrated together to detect and reduce cyber risks associated with the modern cloud based infrastructure. This paper focused on experimenting mainly with four supervised machine learning algorithm to predict the normal and abnormal network traffic based on the features extracted from well-known UNSW-NB15 dataset. The machine learning algorithms that are experimented with in this paper are Naive Bayes, AdaBoost, K- Nearest Neighbors and Random Forest. This paper demonstrates the effectiveness of machine learning and cloud services in protecting and analyzing networks in today's cybersecurity world. The AWS manages cloud services are used to create data analytic solutions to perform, analyze, and visualize network traffics from the UNSW-NB15 dataset. The integrated cloud services and machine learning approach demonstrated in this paper is beneficial to visualize network threats and take proactive approach to reduce those threats.